\documentclass[]{article}
\usepackage{authblk}
\usepackage{fullpage}

\usepackage{multirow}
\usepackage{caption}
\usepackage[english]{babel}
\usepackage[latin1]{inputenc}
\usepackage{amsfonts}
\usepackage{amsmath}
\usepackage{latexsym}
\usepackage{hyperref}
\usepackage{mathrsfs}
\usepackage{etoolbox}
\usepackage{enumitem}
\usepackage{soul}
\usepackage{stackengine}

\usepackage{xcolor}
\usepackage{times}
\usepackage{multirow}
\usepackage{latexsym}

\usepackage{subfig}
\usepackage{multirow}
\usepackage{diagbox}
\usepackage{Definitions}
\newcommand{\xdownarrow}[1]{{\left\downarrow\vbox to #1{}\right.\kern-\nulldelimiterspace}}
\usepackage{algorithm,algorithmic}
\RequirePackage{amsmath,amssymb}
\RequirePackage{colortbl,slashbox}
\RequirePackage[textsize=scriptsize,backgroundcolor=yellow!20,disable]{todonotes} %
\RequirePackage{enumitem} \setlist[description]{itemsep=.3ex}
\RequirePackage{titlesec}
\RequirePackage[skip=.3ex minus.1ex]{caption}
\makeatletter
 \g@addto@macro{\normalsize}{%
 \setlength{\abovedisplayskip}{0pt}
 \setlength{\abovedisplayshortskip}{0pt}
 \setlength{\belowdisplayskip}{0pt}
 \setlength{\belowdisplayshortskip}{0pt}
 \textfloatsep 2ex minus 1ex
 \floatsep \textfloatsep
 \intextsep 1ex}
\makeatother
\definecolor{LightBlue}{rgb}{0.90,0.98,1}
\definecolor{LightOrange}{rgb}{1,0.84,0.80}
\definecolor{LightRed}{rgb}{1,0.50,0.50}
\definecolor{LightGrey}{rgb}{0.9,0.9,0.9}
\definecolor{Green}{rgb}{0.55,0.70,0.0}
\definecolor{LightGreen}{rgb}{0.72,0.91,0.80}
\definecolor{LightYellow}{rgb}{0.94,0.98,0.85}

\newcommand{\best}[1]{\cellcolor{LightGreen}{#1}}
\newcommand{\secBest}[1]{\cellcolor{LightBlue}{#1}}
\newcommand{\better}[1]{\cellcolor{LightBlue}{#1}}

\newcommand{\bidisha}[1]{\textcolor{blue}{[\small{#1}-Bidisha]}}
\newcommand{\bidishadone}[1]{\textcolor{black}{{#1}}}
\newcommand{\our}{CSGen}
\DeclareMathOperator{\idf}{IDF}

\begin{document}
\title{Improved Sentiment Detection via Label Transfer \\
from Monolingual to Synthetic Code-Switched Text}

\author[1]{Bidisha Samanta}
\author[1]{Niloy Ganguly}
\author[2]{Soumen Chakrabarti}

\affil[1]{Indian Institute of Technology, Kharagpur, bidisha@iitkgp.ac.in, niloy@cse.iitkgp.ernet.in}
\affil[2]{Indian Institute of Technology, Bombay, soumen@cse.iitb.ac.in}
\date{}
\maketitle

\begin{abstract}
Multilingual writers and speakers often alternate between two languages in a single discourse, a practice called ``code-switching''. Existing sentiment detection methods are usually trained on sentiment-labeled monolingual text. Manually labeled code-switched text, especially involving minority languages, is extremely rare. Consequently, the best monolingual methods perform relatively poorly on code-switched text. We present an effective technique for synthesizing labeled code-switched text from labeled monolingual text, which is more readily available. The idea is to replace carefully selected subtrees of constituency parses of sentences in the resource-rich language with suitable token spans selected from automatic translations to the resource-poor language. By augmenting scarce human-labeled code-switched text with plentiful synthetic code-switched text, we achieve significant improvements in sentiment labeling accuracy (1.5\%, 5.11\%, 7.20\%) for three different language pairs (English-Hindi, English-Spanish and English-Bengali). We also get significant gains for hate speech detection: 4\% improvement using only synthetic text and 6\% if augmented with real text.
\end{abstract}

\section{Introduction}
\label{sec:Intro}

\if{0}
Code switching is very common on social media platforms like Twitter and Facebook. The simplest type of switching occurs when people alternate between two languages. Code switching is widely used by bilingual users. E.g., a bilingual Spanish/English speaker \todo{political or product tweet?} produced this tweet: ``\emph{Las cuatro de la morning y yo todavia sin poder dormir}, im actually tired!'' which translates to ``Four in the morning and I still can not sleep; I'm actually tired!'' Here {\itshape italicized segments} are in Spanish.
\fi
Sentiment analysis on social media is critical for commerce and governance. Multilingual social media users often use code-switching, particularly to express emotion \cite{rudra2016understanding}. 
However, a basic requirement to train any sentiment analysis (SA) system is the availability of large sentiment-labeled corpora. These are extremely challenging to obtain \cite{chittaranjan2014word, vyas:emnlp2014, barman2014code}, requiring volunteers fluent in multiple languages.


We present \textbf{\our}, a system which provides supervised SA algorithms with 
synthesized unlimited sentiment-tagged code-switched text, without involving human labelers of code-switched text, or any linguistic theory or grammar for code-switching. These texts can then train state-of-the-art SA algorithms which, until now, primarily worked with monolingual text. 

A common scenario in code-switching is that a resource-rich \emph{source} language is mixed with a resource-poor \emph{target} language. Given a sentiment-labeled source corpus, we first create a parallel corpus by translating to the target language, using a standard translator. Although existing neural machine translators (NMTs) can translate a complete source sentence to a target sentence with good quality, it is difficult to translate only designated source segments in isolation because of missing context and lack of coherent semantics.

Among our key contributions is a suite of approaches to automatic segment conversion. Broadly, given a source segment selected for code-switching, we propose intuitive ways to select a corresponding segment from the target sentence, based on maximum similarity or minimum dissimilarity with the source segment, so that the segment blends naturally in the outer source context. Finally, the generated synthetic sentence is tagged with the same sentiment label as the source sentence. The source segment to replace is carefully chosen based on an observation that, apart from natural switching points dictated by syntax, there is a propensity to code-switch between highly opinionated segments.


Extensive experiments show that augmenting scarce natural labeled code-switched text with plentiful synthetic text associated with `borrowed' source labels enriches the feature space, enhances its coverage, and improves sentiment detection accuracy, compared to using only natural text. On four natural corpora having gold sentiment tags, we demonstrate that adding synthetic text can improve accuracy by 5.11\% in English-Spanish, 7.20\% in English-Bengali and (1.5\%, 0.97\%) in English-Hindi (Twitter, Facebook). The synthetic code-switch text, even when used by itself to train SA, performs almost as well as natural text in several cases. Hate speech is an extreme emotion expressed often on social media. On an English-Hindi gold-tagged hate speech benchmark, we achieve 6\% absolute F1 improvement with data augmentation, partly because synthetic text mitigates label imbalance present in scarce real text.


\section{Related Work}
\label{sec:Rel}

Recent SA systems are trained on labeled text \cite{sharma2015text, vilares2015sentiment, joshi2016towards}. For European and Indian code-switched sentiment analysis, several shared tasks have been initiated \cite{barman2014code, SemEval:2017:task4, patra2018sentiment, sequiera:2015, solorio2014overview}. Some of these involve human annotations on code-switched text. \cite{vilares2015sentiment} have annotated the data set released for POS tagging by \cite{solorio2008pos}. \cite{joshi2016towards} had Hindi-English code-switched Facebook text manually annotated and developed a deep model for supervised prediction.

In a different direction, synthetic monolingual text has been created by Generative Adversarial Networks (GAN) \cite{kannan2017adversarial, zhang2016generating, zhang2017adversarial, maqsud2015synthetic}, or Variational Auto Encoders (VAE) \cite{bowman2015generating}. Some of these models can be used to generate sentiment-tagged synthetic text. However, most of them are not directly suitable for generating bilingual code-mixed text, due to the unavailability of sufficient volume of gold-tagged code-mixed text.
~\cite{VACSIJCAI} proposed a generative method using a handful of gold-tagged data; but they cannot produce sentence level tags.
Recently, \cite{Pratapa:2018:ACL} used linguistic constraints arising from Equivalence Constraint Theory to design a code-switching grammar that guides text synthesis. Earlier, \cite{bhat2016grammatical} presented similar ideas, but without empirical results. In contrast, \our{} uses a data-driven combination of word alignment weights, similarity of word embeddings between source and target, and attention \cite{bahdanau2014neural}.

\section{Generation of code-switched text}

\our{} takes a sentiment-labeled source sentence $s$ and translates it into a target language sentence~$t$. Then it generates text with language switches on particular constituent boundaries. This involves two sub-steps: select a segment in $s$ (\S\ref{sec:source-selection}), and then select text from $t$ that can replace it~(\S\ref{sec:target-replacement}--\S\ref{sec:projection}). 
This generation process is sketched in Algorithm~\ref{alg:main}.


\subsection{Sentiment-oriented source segment selection}
\label{sec:source-selection}

In this step, our goal is to select a contiguous segment from the source sentence that could potentially be replaced by some segment in the target sentence. (Allowing non-contiguous target segments usually led to unnatural sentences.) Code switching tends to occur at constituent boundaries \cite{SankoffPoplack1981}, an observation that holds even for social media texts~\cite{begum2016functions}. Therefore, we apply a constituency parser to the source sentence. Specifically, we use the Stanford CoreNLP shift-reduce parser \cite{zhu2013fast} to generate a parse tree\footnote{\url{http://stanfordnlp.github.io/CoreNLP/}}. Then we select segments under non-terminals, i.e., subtrees, having certain properties, chosen using heuristics informed by patterns observed in real code-switched text.

\paragraph{NP and VP:} 
We allow as candidates all subtrees rooted at NP (noun phrase) and VP (verb phrase) nonterminals, which may cover multiple words. Translating single-word spans is more likely to result in ungrammatical output~\cite{SankoffPoplack1981}. 

\paragraph{SBAR:} 
Bilingual writers often use a clause to provide a sentiment-neutral part and then switch to another language in another sentence-piece to express an opinion or vice-versa. An example is ``Ramdhanu ended with tears \textit{kintu sesh ta besh onho rokom etar}'' (Ramdhanu ended with tears but the ending was quite different). Here the constituent ``\textit{but the ending was quite different}" comes under the subtree of SBAR.


\begin{algorithm}[th]
\caption{\our{} overview.}
\label{alg:main}
\begin{algorithmic}[1]
	\STATE \textbf{Input:} Sentiment-labeled source sentences $S=\{(s_n, y_n)\}$
	\STATE \textbf{Output:} Synthetic code-switched sentences $C=\{(c, y)\}$
	\STATE $t_n \leftarrow \text{Translate}(s_n)\; \forall s_n\!\in\! S$ \texttt{/* Make parallel corpus */}
	\STATE $C \leftarrow \varnothing$ 
	\FOR{each parallel sentence pair $s, t$}
 	\STATE \texttt{/*Collect word alignment signals*/}
 	\STATE $a \leftarrow \text{AttentionScore}(s,t),\; g \leftarrow \text{GizaScore}(s,t)$
 	\STATE \texttt{/* Source segment selection */}
 	\STATE $P \leftarrow \text{SentimentOrientedSegmentSelection}(s)$
 	\FOR{each segment $p_s \in P$ to replace}
 	\STATE \texttt{/* Target segment selection */}
 	\STATE $\hat{q}^1, \hat{q}^2 \leftarrow \text{MaxSimTargetSeg}(s, p_s, t, a, g)$
 	\STATE $\hat{q}^3\!, \hat{q}^4 \!\!\!\leftarrow\!\!\!\text{MinDissimTargetSeg}(s, p_s, t, 1\!-\!a, 1\!-\!g)$
 	\STATE \texttt{/*Code-switched text generation*/}
 	\STATE $C_k \leftarrow \text{Project}(s,t,p_s,\hat{q}^k)$ where $k\in\{1,\ldots,4\}$
 	\STATE $C \leftarrow C \cup \text{SelectBest}(\{C_k: k\in\{1,\ldots,4\}\})$
 	\ENDFOR
	\ENDFOR
	\STATE $C \leftarrow \text{Threshold}(C)$ \texttt{/* Retain only best replacements */}
	\end{algorithmic}
\end{algorithm}

\paragraph{Highly opinionated segments:}
We also include segments which have a strong opinion polarity, as detected by a (monolingual) sentiment analyzer \cite{gilbert2014vader}. 
E.g., the tweet ``\textit{asimit khusi prasangsakako ke beech} \dots\ as India won the world cup after 28 years'' translates to ``Unlimited happiness among fans \dots\ as India won the world cup after 28 years''. 

\begin{figure}[t]
\centering\includegraphics[clip, width=0.5\hsize, ]{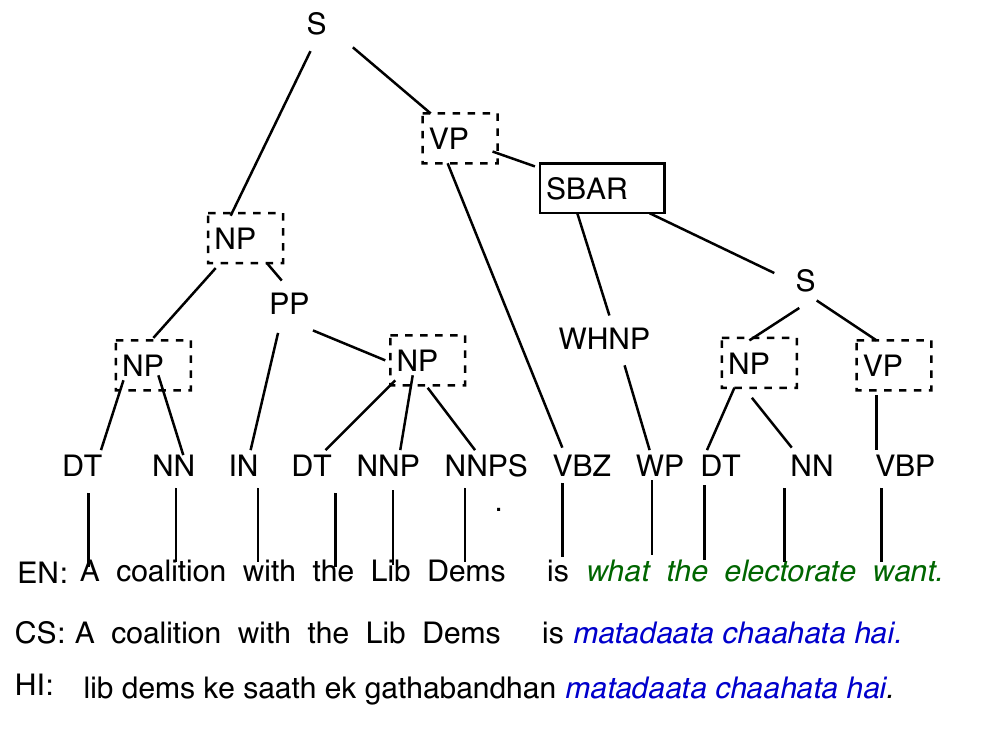}
\caption{A phrase-structure tree for a sample synthesis. Dotted-boxes around constituents indicate that they are candidates for replacement on the source side~(\S\ref{sec:source-selection}). EN:~English source sentence, HI:~Hindi target sentence, CS:~code-switched sentence. The \textcolor{blue}{\textit{italicized segment}} is the target segment to replace the \textcolor{green!40!black}{\textit{source segment}} under the non-terminal SBAR.}
\label{fig:constituent_tree}
\end{figure}

An example sentence, its parse tree, and its candidate replacement segments are shown in Figure~\ref{fig:constituent_tree}. In Algorithm~\ref{alg:main}, $p_s \in P$ denotes the set of candidate replacement subtrees, which correspond to segments. For each candidate segment, we generate a code-switched version of the source sentence, as described next.

\subsection{Target segment selection}
\label{sec:target-replacement}

Given a source sentence $s$, corresponding target $t$, and one (contiguous) source segment $p_s = \{w_s^i \cdots w_s^{i+x}\}$, the goal is to identify
the best possible a contiguous target segment $q_t=\{w_t^j \cdots w_t^{j+y}\}$ that could be used to replace $p_s$ to create a realistic code-switched sentence. We adopt two approaches to achieve this goal: (a)~selecting a target segment that has maximum similarity with $p_s$, and (b)~selecting a target segment having minimum dissimilarity with $p_s$, for various definitions of similarity and dissimilarity. Below, we describe methods that achieve this goal after describing several alignment scores which will be used in these methods. Overall, these lead to target segments $\hat{q}^1_t, \hat{q}^2_t, \ldots$ shown in Algorithm~\ref{alg:main}, with $t$ removed for clarity.

\subsubsection{\bfseries Word alignment signals}
\label{sec:corres}
Signals based on word alignment methods are part of the recipe in choosing the best possible $q_t$ given the sentence pair and $p_s$.

\paragraph{\bfseries GIZA score:}
The standard machine translation word alignment tool Giza++ ~\cite{och03:asc} uses IBM statistical word alignment models~1--5 \cite{fernandez2008improving, schoenemann2010computing, brown1993mathematics, riley2012improving}. This tool incorporates principled probabilistic formulations of the IBM models and gives a correspondence score $G[w_t^i,w_s^j]$ between target and source words for a given sentence pair. This word-pair score is used as a signal to find the best~$\hat{q}_t$.

\paragraph{\bfseries NMT attention score:} 
Given an attention-guided trained seq\-uence-to-sequence neural machine translation (NMT) model \cite{bahdanau2014neural, luong2015effective} and sentence pair $s, t$, we use the attention score matrix $A[w_t^i,w_s^j]$ as an alignment signal.

\paragraph{\bfseries Inverse document frequency (rarity):}
The inverse document frequency ($\idf$) of a word in a corpus signifies its importance in the sentence~\cite{Rijsbergen79}. We define $\Ical(w) = \sigma( a\, \idf(w) - b)$ as a shifted, squashed $\idf$ that normalizes the raw corpus-level score. Here $\sigma$ is the sigmoid function and parameters $a$ and $b$ are empirically tuned. This IDF-based signal is optionally incorporated while choosing~$\hat{q}_t$.

\subsubsection{\bfseries Target segment with maximum similarity}
\label{sec:phrasal-alignment}

Given word-pair scores derived from either Giza++ or NMT attention described in \S\ref{sec:corres}, we formulate two methods for identifying target segments. First, we identify the best target segment given Giza++ scores, $G[\cdot,\cdot]$, as follows:
\begin{align}
\hat{q}^1_t \leftarrow \argmax_{q_t} \prod_{w_t \in q_t} \sum_{w_s \in p_s} G[w_t,w_s]
\end{align}
For each word in $q_t$, we compute the total attention score concentrated in $p_s$ and then multiply them as if they are independent.

Second, we use the attention score learned by the NMT system of \cite{luong17} (a bidirectional LSTM model with attention). Essentially, given the attention score $A[\cdot,\cdot]$ between target and source words, we intend to select the target segment $q_t$ whose maximum attention is concentrated in the given~$p_s$.

Initial exploration of the above method revealed that the attention of a target word may spread out over several related but less appropriate source words, and accrue better overall similarity than a single more appropriate word. Here IDF can come to the rescue, the intuition being that words $w_t^i$ and $w_s^j$ with very different IDFs are less likely to align, because (barring polysemy and synonymy) rare (common) words in one language tend to translate to rare (common) words in another. This intuition is embodied in the improved formulation:
\begin{align}
\hat{q}_{t}^{2} \leftarrow \argmax_{q_t} \prod_{w_t \in q_t} \Ical(w_t) 
\sum_{w_s \in p_s} \Ical(w_s) A[w_t,w_s]
\end{align}
Informally, if a source segment contains many rare words, the target segment should also have a similar number of rare words from the target domain, and vice-versa.

\subsubsection{\bfseries Target segment with minimum dissimilarity}
\label{sec:emd}

We examine an alternative method for identifying target segments that leverage the \textit{Earth Mover's Distance (EMD)} \cite{vaserstein:emd1969}. \cite{kusner2015word} extended EMD to the \textit{Word Mover Distance} to measure the dissimilarity between documents by `transporting' word vectors from one document to the vectors of the other. In the same spirit, we define a dissimilarity measure between $p_s$ and candidate target segments using EMD. We present here EMD as a minimization over fractional transportation matrix $\Fb \in \RR^{|q_t| \times |p_s|}$ as below:
\begin{align}
\text{EMD}(q_t, p_s) &= \min_{\Fb} \sum_{i=1}^{|q_t|} \sum_{j=1}^{|p_s|} F_{i,j} d_{i,j}
\label{eq:emd}
\end{align}
where $\sum_i{F_{i,j}} = \frac{1}{|{q_t}|}$ and $\sum_j{F_{i,j}} = \frac{1}{|{p_s}|}$ and 
$d_{i,j}$ is a distance metric between a target and a source word pair, given suitable representations.
Finally, we choose the target segment which is least dissimilar to a given source segment defined by the EMD.
We compute $d_{i,j}$ in two ways, described below.

\paragraph{\bfseries Attention-based distance:}
Here the distance between the embeddings is defined as:
\begin{align}\label{eq:attn}
d_{i,j}^A = 1 - A[w_t^i,w_s^j]
\end{align}

\paragraph{\bfseries Giza-based distance:}
Similarly we can compute the distance using Giza score as:
\begin{align}\label{eq:giza}
d_{i,j}^G = 1 - G[w_t^i,w_s^j]
\end{align}
Given the two types of distances in Eq.~\eqref{eq:attn}--\eqref{eq:giza} and the definition of EMD in Eq.~\eqref{eq:emd} we can formulate two methods for identifying target segments:
\begin{align}
\hat{q}_{t}^{k} \leftarrow \argmin_{q_t} \min_{\Fb} \sum_{i=1}^{|q_t|} \sum_{j=1}^{|p_s|} F_{i,j} d_{i,j}^k 
\end{align}
where $k \in \{3,4\}$ and $d_{i,j}^3\equiv d_{i,j}^A$ and $d_{i,j}^4\equiv d_{i,j}^G$.

We can also use Euclidean distance as $d_{i,j}$. However, this method requires multilingual word embeddings for every word to calculate the distance. The volume of labeled source text we can use is usually smaller than the vocabulary size, making it difficult to learn reliable word embeddings. Also, if these corpora contain informal social media text like the ones described in \S \ref{sec:source}, then publicly available pretrained word embeddings exclude a significant percentage of them.

\subsection{Projecting target segments}
\label{sec:projection}

Given a source sentence $s$ with designated segment $p_s$ to replace, and target sentence $t$, we have by now identified four possible target segments $\hat{q}_{t}^{k}$ where 
$k \in \{1, \ldots, 4\}$ as described in~\S\ref{sec:phrasal-alignment}--\S\ref{sec:emd}. We now \emph{project} the target segment to the source sentence, meaning, (a)~replace the source segment with the target segment and (b)~transliterate the replacement using the Google Transliteration API to the source script\footnote{\protect \path{http://www.google.com/transliterate?langpair=hi|en&text=<text>}}. This creates four possible synthetic code-switched sentences for each instance of $(s,p_s,t)$.
\bidishadone{Finally, we transfer the labels of the original monolingual corpora to the generated synthetic text corpora.}

\subsection{Best candidate via reverse translation}
\label{sec:reverse}
From these four code-switched sentences $c_1,\ldots,c_4$, we wish to retain the one that retains most of the syntactic structures of the source sentence. Each code-switched sentence $c_k$ has an associated score as defined in \S\ref{sec:target-replacement}. We use two empirically tuned thresholds: a lower cut-off for the similarity score of $c_1, c_2$ and an upper cut-off for the dissimilarity score of $c_3, c_4$, to improve the quality of candidates retained. These scores are not normalized and cannot be compared across different methods. Therefore, we perform a reverse translation of each candidate back to the source language using the Google translation API to obtain~$\tilde{s}$. We retain the candidate whose retranslated version $\tilde{s}$ has the highest BLEU score \cite{papineni2002bleu} wrt~$s$. In case of a tie, we select the candidate with maximum word overlap with~$s$.

\subsection{Thresholding and stratified sampling}

In addition to retaining only the best among code-switched candidates $c_{1,\ldots,4}$, we discard the winner if its BLEU score is below a tuned threshold. Further, we sample source sentences such that the surviving populations of sentiment labels of the code-switched sentences match the populations in the low-resource evaluation corpus. Another tuned system parameter is the amount of synthetic text to generate to supplement the gold text. 

We do not depend on any domain coherence between the source corpus used to synthesize text and the gold `payload' corpus --- this is the more realistic situation. Our expectation, therefore, is that adding some amount of synthetic text should improve sentiment prediction, but excessive amounts of off-domain synthetic text may hurt it. In our experiments we grid search the synthetic:gold ratio between 1/4 and 2 using 3-fold cross validation.

\section{Experiments}
\label{sec:exp}

We demonstrate the effectiveness of augmenting gold code-switched text with synthetic code-switched text. We also measure the usefulness of synthetic text without gold text. In this section, we will first describe the data sets used to generate the synthetic text and then the resource-poor labeled code-switched text used for evaluation. Next, we will present the method used for sentiment detection, baseline performance, and finally our performance, along with a detailed comparative analysis.

\subsection{Source corpora for text synthesis}
\label{sec:source}

We use publicly available monolingual sentiment-tagged (positive, negative or neutral) gold corpora in the source language.
\begin{description}[leftmargin=*]

\item[ACL:]
\cite{dong2014adaptive} released about 6000 manually labeled English tweets.

\item[Election:]
\cite{wang2017tdparse} published about 5000 human-labeled English tweets.

\item[Mukherjee:]
This data set contains about 8000 human-labeled English tweets~\cite{mukherjee2012sentiment, mukherjee2012twisent}.

\item[Semeval shared task:]
This provides about 10000 human-labeled English tweets~\cite{SemEval:2017:task4}.

\item[Union:]
This is the union of above mentioned different data sets.

\item[Hatespeech:]
We collected 15K tagged English tweets from ~\cite{founta2018large} which consists of 4.7K \textit{abusive}, 1.7K \textit{hateful} and 4K \textit{normal} tweets.
\end{description}


We picked Spanish, which is homologous to English, and Hindi and Bengali, which are comparatively dissimilar to English, for our experiments. We translated these monolingual tweets to Spanish, Hindi and Bengali using Google Translation API\footnote{\url{https://translation.googleapis.com}} and used as parallel corpus to train attention-based NMT models and statistical MT model (GIZA) to learn the word alignment signals as described in~\S\ref{sec:corres}.

\subsection{Preliminary qualitative analysis}

Analysis of texts synthesized by various mechanisms proposed in \S\ref{sec:target-replacement} shows that similarity based methods contribute 82--85\% of the best candidates and the rest come from dissimilarity based methods. Similarity-based methods using NMT attention and Giza perform well because the segments selected for replacement often constitute nouns and entity mentions, which have a very strong alignment in the corresponding target segment. NMT attention and Giza-EMD perform well when segments contract or expand in translation.

\subsection{Low-resource evaluation corpora}
To evaluate the usefulness of the generated synthetic tagged sentences as a training set for sentiment analysis, we have used three different code-switched language pair data sets. Each data set below was divided into 70\% training, 10\% validation and 20\% testing folds. 
The training fold was (or was not) augmented with synthetic labeled text to train sentiment classifiers, which were then applied on the test fold judge the quality of synthesis.



\begin{description}[leftmargin=*]

\item[HI-EN, FB (Hindi-English, Facebook):]
\cite{joshi2016towards} released around 4000 labeled code-switched sentences from the Facebook timeline of Narendra Modi (Indian Prime Minister) and Salman Khan (Bollywood actor).

\item[HI-EN, TW (Hindi-English, Twitter):]
This is a shared task from ICON 2017 \cite{patra2018sentiment} with 15575 instances.

\item[ES-EN (English-Spanish):]
We collected 2883 labeled tweets specified by \cite{vilares2015sentiment}.

\item[BN-EN (Bengali-English):]
This is another shared task from ICON 2017 \cite{patra2018sentiment} with 2499 instances.

\item[HI-EN, Hatespeech:]
\cite{bohra2018dataset} published 4000 manually-labeled code-switched Hindi-English tweets: 1500 exhibiting \textit{hate speech} and 2500 \textit{normal}. We also found a significant number of abusive tweets marked \textit{hate speech}. For uniformity, we merged hate speech tweets and abusive tweets.

\end{description}

\subsection{Sentiment classifier}

We adopt the sub-word-LSTM system of \cite{joshi2016towards}. We prefer this over feature-based methods because (a)~feature extraction for code-switched text is very difficult, and varies widely across language pairs, and (b)~the vocabulary is large and informal, with many tokens outside standard (full-) word embedding vocabularies and (c)~sub-word-LSTM captures semantic features via convolution and pooling.


\paragraph{\bfseries Loss functions:}
If the sentiment labels $\{-1,0,+1\}$ are regarded as categorical, cross-entropy loss is standard. However, prediction errors between the extreme polarities $\{-1, +1\}$ need to be penalized more than errors between \{-1,0\} or \{0,+1\}. Hence, we use ordinal cross-entropy loss \cite{niu2016ordinal}, introducing a weight factor proportional to the order of intended penalty multiplied with the cross entropy loss. On the test fold, we report 0/1 accuracy and per class micro-averaged F1 score.

\paragraph{\bfseries Baseline and prior art:}
Our baseline scenario is a self-contained train-dev-test split of the gold corpus. The primary prior art is the work of \cite{Pratapa:2018:ACL}. 

\paragraph{\bfseries Feature space coherence:}
Our source corpora are quite unrelated to the gold corpora. Table~\ref{tab:feature} shows that the average Euclidean distance between feature space of gold training and testing texts is much lower than that between gold and synthetic texts. While this may be inescapable in a low-resource situation, the gold baseline does not pay for such decoherence, which can lead to misleading conclusions. 

\begin{table}[ht]
\centering
\begin{tabular}{|l|l|c|c|c|} \hline
& HI\_EN,TW & HI\_EN,FB & ES\_EN & BN\_EN \\ \hline

ACL &2.21 (2.13) &3.72 (2.24) & \cellcolor{red!8} 2.09 (1.73) &4.11 (2.67) 
\\ \hline
Election &2.40 (2.12) & \cellcolor{red!8} 6.27 (2.67) &1.58 (1.49)& 5.23 (2.43)
\\ \hline
Mukherjee & \cellcolor{red!8}2.47 (2.33) &3.82 (2.26) &1.64 (1.64) &5.18 (2.50) 
\\ \hline
Semeval &2.23 (2.11) &4.04 (2.26) &1.69 (1.67) &3.63 (2.59) 
\\ \hline
Union & 2.55 (2.15)& 3.80(2.65) & 1.65 (1.53) & \cellcolor{red!8} 5.48 (2.56)
\\ \hline
\rowcolor{gray!15} Gold &2.05 &1.87 &1.64& 1.83
\\ \hline 
\end{tabular}
\caption{Average pairwise Euclidean distance between training data and test data features. Rows correspond to standalone (respectively, augmented) text for training. Gray: reference distance of gold test from gold train. Red: largest distance observed.}
\label{tab:feature}
\end{table}

\begin{table*}
\resizebox{\hsize}{!}{
\begin{tabular}{|l|l|l|l|l||l|l|l|l|} \hline
\backslashbox{Train}{Test}
& \begin{tabular}[c]{@{}l@{}}HI\_EN\\ (TW)\end{tabular} & \begin{tabular}[c]{@{}l@{}}HI\_EN\\ (FB)\end{tabular} & ES\_EN & BN\_EN &
\begin{tabular}[c]{@{}l@{}}HI\_EN\\ (TW)\end{tabular} & \begin{tabular}[c]{@{}l@{}}HI\_EN\\ (FB)\end{tabular} & ES\_EN & BN\_EN\\ \hline
 & \multicolumn{4}{c||}{Categorical cross entropy training loss} & \multicolumn{4}{c|}{Ordinal cross entropy training loss} \\ \hline

ACL &\better 51.80 (52.59) & 62.59 (65.33) & \best 44.80 (48.84) & \better 52.59 (59.81) &\better 52.68 (53.76) & \better 62.72 (64.22) & \best 45.69 (50.31) & \better 50.08 (51.60)\\ \hline

Election & \better 52.84 (54.59) & \best 65.59 (66.80) & \better 43.07 (43.07) & \secBest 57.99 (59.00) & \best52.89 (54.64) &\secBest 64.88 (65.26) & \better 45.21 (45.80) & \secBest 53.40 (55.60)\\ \hline
Mukherjee &\best 53.76 (54.69) & \better 64.82 (65.85) & \better 47.06 (46.36) & \better 49.79 (57.40)
&\secBest 53.79 (53.53) &\better 59.66 (64.57) & 44.85 (43.07) &\better 51.00 (49.40) 
\\ \hline
Semeval & \better 52.99 (54.19) & \better 65.33 (65.46) & \better 47.36 (44.69) & \best 53.40 (59.99) & \better 52.99 (53.83) &\better 63.26 (64.66) & \better 47.36 (44.64) & \secBest 53.14 (57.59)\\ \hline
Union &\better 53.28 (54.64) & \better 65.50 (65.99) & \better 44.24 (45.32) &\secBest 57.23 (59.89)&\better 53.65 (53.69)& \best 64.30 (67.65) &\secBest 44.04 (46.00) &\best 53.40 (57.40)\\
\hline
MSR &\secBest 54.50 & \better 65.58 &\secBest 48.14 &\secBest 59.79 & \better 53.69 & 62.80 &\secBest 47.50 & \better 52.8 \\ \hline
\rowcolor{gray!15} Gold & 52.26 & 65.37 & 42.6 & 55.19& 52.34 & 64.29 & 45.20 & 50.39 \\ \hline
\end{tabular}
}
\caption{Accuracy (\%) on 20\% test data after training with augmented and only gold text. Rows correspond to sources of augmentation. In most cells we show (A)~no thresholding or stratification and (B)~with thresholding and stratification (within brackets). Gray: reference accuracy with only gold training. Blue: A or B or MSR outperforms gold. 
Green: B performs best. Row `MSR' uses text synthesized by \cite{Pratapa:2018:ACL}.}
\label{tab:augmented}
\end{table*}

\paragraph{\bfseries Training regimes:}
Absence of coherent tagged gold text may lead to substantial performance loss. Hence, along with demonstrating the usefulness of augmenting natural with synthetic text, we also measure the efficacy of synthetic text on its own. We train the SA classifier with three labeled corpora: (a)~limited gold code-switched text, (b)~gold code-switched text augmented with synthetic text and (c)~only synthetic text. Then we evaluate the resulting models on labeled gold code-switched test fold.

\subsection{Sentiment detection accuracy}
\label{sec:acc}

Table~\ref{tab:augmented} shows the benefits of augmenting natural with synthetic text.
Test accuracy increases further (shown in brackets) if thresholding and stratified sampling are used.
Gains for HI\_EN,TW, HI\_EN,FB, ES\_EN and BN\_EN are 
1.5\% (2.43\%), 0.23\% (1.43\%), 4.76\% (6.24\%), and 
2.8\% (4.8\%) 
respectively. Categorical cross-entropy loss 
was used here. 
Similar improvements in accuracy of 
1.45\% (1.5\%), 0.59\% (0.97\%), 2.16\% (5.11\%), 3\% (7.20\%) 
are observed after training with ordinal loss function. 
Our conclusion is that \emph{careful augmentation with synthetic data can lead to useful gain in accuracy}. Moreover, by selecting synthetic text which is syntactically more natural, even larger gains can be achieved. Notably, the distance between training and test features (Table~\ref{tab:feature}) is negatively correlated with accuracy gain (Pearson correlation coefficient of $-0.48$).

\paragraph{\bfseries Comparison with \cite{Pratapa:2018:ACL}: }
They depend on finding correspondences between constituency parses of the source and target sentences. However, the common case is that a constituency parser is unavailable or ineffective for the target language, particularly for informal social media. They are thus restricted to synthesizing text from only a subset of monolingual data. Training SA with natural text augmented with their synthesized text leads to poorer accuracy, albeit by a small amount, than using \our{}. The performance is worse for target languages that are more resource-poor.

\paragraph{\bfseries Ordinal vs.\ categorical loss:}
Table~\ref{tab:augmented} shows that ordinal loss helps when the neutral label dominates. However, neither is a clear winner and the gains are small. Therefore, we use categorical loss henceforth.

\paragraph{\bfseries Choice of monolingual corpus:}
Across all monolingual corpora, \emph{Election} performs consistently well. Best test performance on HI\_EN,TW was obtained by synthesizing from the \emph{Mukherjee} corpus. Text synthesized from \emph{Election} provides the best results for HI\_EN,FB for both setups. The performance of \emph{Union} is also good but not the best. This is because although a larger and diverse amount of data is available which ensures its quality, the Euclidean distance between test data and some individual corpora is still large.


\begin{table}[th]
\centering
\small
\begin{small}
\begin{tabular}{|l|l|l|l|l|l|l|}
\hline
& \multicolumn{3}{c|}{\begin{tabular}[c]{@{}l@{}}Categorical Cross \\Entropy training\end{tabular}
 } & \multicolumn{3}{c|}{
 \begin{tabular}[c]{@{}l@{}}Ordinal Cross \\Entropy training\end{tabular}} \\
 \hline
 & Pos & Neu & Neg & Pos & Neu & Neg \\
 \hline
 & \multicolumn{6}{c|}{HI\_EN,TW} 
 \\
 \hline
 \hline
 \our{} & \better 0.52 & 0.62 & \better 0.38& \better 0.55 & \better 0.63 & 0.34 
 \\
 \hline
Gold & 0.48 & 0.63 & 0.24 & 0.50 & 0.62 & 0.35
\\
\hline
 & \multicolumn{6}{c|}{HI\_EN,FB} 
 \\
 \hline
 \hline
 \our{} 
 & 0.59 & 0.73 &\better 0.56 & \better 0.62 & 0.71 & \better 0.55
 \\ 
 \hline
Gold 
& 0.60 & 0.74 & 0.54 & 0.60 & 0.71 & 0.44 \\
 \hline
 & \multicolumn{6}{c|}{ES\_EN} \\
 \hline
 \hline
 \our{} 
 & 0.38 & \better 0.53 & 0.37 & \better 0.48 & 0.50 & 0.42
 \\ 
 \hline
Gold 
& 0.47 & 0.44 & 0.41 & 0.40 & 0.53 & 0.43
 \\ 
 \hline
 & \multicolumn{6}{c|}{BN\_EN} \\ \hline
 \hline
 \our{} 
 & \better 0.63 & 0.49 & 0.58 & \better 0.55 & 0.47 & 0.59
 \\ 
 \hline
Gold 
& 0.55 & 0.51 & 0.65 & 0.37 & 0.49 & 0.61
 \\ \hline
\end{tabular}
\end{small}
\caption{F1 score for each 
class prediction. Blue: \our{} is better than Gold.}
\label{tab:prec_rec}
\end{table}

\subsection{Sentiment detection F1 score}

Beyond 0/1 accuracy, Table~\ref{tab:prec_rec} shows F1 score gains. Election yields consistently good results. We have reported the F1 score gain for different sentiment classes only for Election in Table~\ref{tab:prec_rec} for brevity. Augmenting synthetic data with gold data yields better F1 score than training only with gold tagged data. Also, it is interesting to observe that there is a sharp drop of F1 score for HI\_EN,FB and BN\_EN data sets for Gold data while training with ordinal cross entropy function across all the sentiment labels. As described in \S\ref{sec:acc}, this is due to non-discriminative features. However, mixing them with synthetic data helps in achieving better results.

\begin{table}[th]
\small
\centering
\begin{tabular}{|l|r|r|r|r|} \hline
\backslashbox{Train}{Test}
& \begin{tabular}[c]{@{}l@{}}HI\_EN\\ (TW)\end{tabular} & \begin{tabular}[c]{@{}l@{}}HI\_EN\\ (FB)\end{tabular} & ES\_EN & BN\_EN \\ \hline

ACL & 40.33 & 49.96 & 38.40 & \better 47.81 \\ \hline
Election & \better 47.22 & 48.78 & 31.20 & 42.44 \\ \hline
Mukherjee & 46.22 & \better 48.98 & 39.76 & 44.42 \\ \hline
Semeval & 45.80 & 48.38 & 39.18 & 45.99 \\ \hline
Union &43.50 &49.80 & \better 41.90 & 41.20 \\
\hline
\rowcolor{gray!15} Gold & 52.26 & 65.37 & 42.60 & 55.19 \\ \hline
\end{tabular}
\caption{Percent accuracy on 20\% test data after training on only synthetic and only gold text. Each row corresponds to a source. Grey: Accuracy achieved with only gold training. Blue: The closest accuracy achieved to best.}
\label{tab:synth}
\end{table}

\begin{table*}[ht]
\scalebox{0.79}{
\begin{tabular}{llll}
\hline
Category of failure & Example sentence & Gold & Predicted \\ \hline
\hline
\begin{tabular}[c]{@{}l@{}}Keywords with different\\ polarity\end{tabular} & \begin{tabular}[c]{@{}l@{}}manana voy conquistar la will forever be an \textcolor{green!60!black}{amazing} song \textcolor{red!70!black}{not} because me \\ la dedicaron but because my momma always jams to it\end{tabular} & Positive & Negative \\
& \textcolor{blue}{\begin{tabular}[c]{@{}l@{}}``tomorrow I will conquer the will" forever be an amazing song not because \\ they dedicated it to me but because my momma always jams to it.\end{tabular}}
& & \\
\cline{2-4}
 & \begin{tabular}[c]{@{}l@{}}twin brothers lost in fair \textcolor{green!60!black}{reunited} in adulthood amidst dramatic circumstances \\ ei themer movie akhon ar viewers der \textcolor{green!60!black}{attract} kore \textcolor{red!70!black}{na}\end{tabular} & Neutral & Positive \\
 & \textcolor{blue} {\begin{tabular}[c]{@{}l@{}}twin brothers lost in fair reunited in adulthood amidst dramatic circumstances,\\ this theme does not yet attract viewers.\end{tabular}} & & \\
 \hline
\begin{tabular}[c]{@{}l@{}}Ambiguous overall \\ meaning\end{tabular} & \begin{tabular}[c]{@{}l@{}}elizaibq ellen quiere entrevistar julianna margulies clooney says she is tough \\ cookie she is hard one to crack\end{tabular} & Negative & Neutral \\
&\textcolor{blue}{ \begin{tabular}[c]{@{}l@{}} elizaibq ellen wants to interview julianna margulies clooney says she is tough\\ cookie she is hard one to crack \end{tabular}} & & \\
\cline{2-4}
 & hum kam se kam fight ker haaray lekin tum loog zillat ki maut maaray gaye & Positive & Negative \\
 & \textcolor{blue}{We lost at least after a fight, but you died a terrible death.}
& & \\

\hline
\end{tabular}
}
\caption{Examples cases of failure in prediction. Red: Negative Polarity words. Green: Positive polarity words. Blue represents the English translation of the code-switched sentence.}

\label{tab:error}
\end{table*}

\subsection{Performance of standalone synthetic data}

The accuracy of using only synthetic data as training is reported in Table~\ref{tab:synth}. We can see that for EN\_HI,TW and EN\_ES the synthetic data is very close to the gold data performance (lagging by 5.04\% and 2.84\%).
However, it performed poorly for HI\_EN,FB and BN\_EN dataset. This is because there is heavy mismatch between the synthetic text set generated and the test data distribution (Table~\ref{tab:feature}) in these two datasets. 
The Pearson rank correlation coefficient between the distance (between test and training set) measures and relative accuracy gain is highly negative,~$-0.66$.

To further establish the importance of domain coherence, we report on an experiment performed with HI\_EN,FB gold dataset. This dataset has texts corresponding to two different entities namely {\em Narendra Modi} and {\em Salman Khan}. Training SA with natural text corresponding to one entity and testing on the rest leads to a steep accuracy drop from 65.37\% to 52.32\%.

\subsection{Error analysis}

\todo{no space for EN translations?}
We found two dominant error modes where synthetic augmentation confuses the system.
Table~\ref{tab:error} shows a few examples.
The first error mode can be triggered by the presence of words of different polarities, one polarity more common than the other, and the gold label being the minority polarity. The second error mode is prevalent when the emotion is weak or mixed. Either there is no strong opinion, or there are two agents, one regarded positively and the other negatively.

\subsection{Hate speech detection results}

Table~\ref{tab:hatespeech} shows hate speech detection results. Training with only synthetic text after thresholding and stratified sampling outperforms training with only gold-tagged text by 4\% F1, and using both gold and synthetic text gives a F1 boost of 6\% beyond using gold alone. Remarkably, synthetic text alone outperforms gold text, because gold text has high class imbalance, leading to poorer prediction. Because we can create arbitrary amounts of synthetic text, we can balance the labels to achieve better prediction.

\begin{table}[!ht]
\small
\centering
\begin{tabular}{|l|l|l|l|}
\hline
 & Prec & Recall & F-score \\ \hline
Only synthetic & \cellcolor{green!10}0.58 (0.63) & 0.60 (0.63) & 0.51 (0.52) \\ \hline
Synthetic +Gold & 0.59 (0.60) & \cellcolor{green!10}0.63 (0.63) & \cellcolor{green!15}0.53 (0.54) \\ \hline
Gold & 0.40 & 0.62 & 0.48 \\ \hline
\end{tabular}
\caption{Hate speech results (3-fold cross val.). In most cells we show performance without thresholding and stratification (within bracket with thresholding and stratification). Green: Best performance in each column.}
\label{tab:hatespeech}
\end{table}

\section{Conclusion}
\label{sec:End}

Code-mixing is an important and rapidly evolving mechanism of expression among multilingual populations on social media. Monolingual sentiment analysis techniques perform poorly on code-mixed text, partly because code-mixed text often involves resource-poor languages. Starting from sentiment-labeled text in resource-rich source languages, we propose an effective method to synthesize labeled code-mixed text without designing switching grammars. Augmenting scarce natural text with synthetic text improves sentiment detection accuracy.

\section*{Acknowledgments}
Bidisha Samanta was supported by Google India Ph.D. Fellowship. We would like to thank Dipanjan Das and Dan Garrette for their valuable inputs. Soumen Chakrabarti was partly supported by IBM.

\bibliography{main}
\bibliographystyle{abbrv}
\end{document}